# Optimizing Retrieval-Augmented Generation (RAG) for Colloquial Cantonese: A LoRA-Based Systematic Review

**David Santandreu Calonge[1], Linda Smail[2]**

[1] Center for Teaching and Learning, Mohamed bin Zayed University of Artificial Intelligence, Abu Dhabi, United Arab Emirates, david.santandreu@mbzuai.ac.ae, https://orcid.org/0000-0003-0101-8758

[2] College of Interdisciplinary Studies, Zayed University, Dubai, United Arab Emirates, Linda.Smail@zu.ac.ae, https://orcid.org/0000-0001-9388-1334

*Corresponding author: Dr. David Santandreu Calonge, Center for Teaching and Learning, Mohamed bin Zayed University of Artificial Intelligence, Abu Dhabi, United Arab Emirates, david.santandreu@mbzuai.ac.ae, https://orcid.org/0000-0003-0101-8758

## Abstract

This review examines recent advances in Parameter-Efficient Fine-Tuning (PEFT), with a focus on Low-Rank Adaptation (LoRA), to optimize Retrieval-Augmented Generation (RAG) systems like Qwen3, DeepSeek, and Kimi. These systems face challenges in understanding and generating authentic Cantonese colloquial expressions due to limited annotated data and linguistic variability. The review evaluates the integration of LoRA within RAG frameworks, benchmarks PEFT methods for retrieval and generation accuracy, identify domain adaptation strategies under limited data, and compares fine-tuning techniques aimed at improving semantic fidelity under data-scarce conditions. A systematic analysis of recent studies employing diverse LoRA variants, synthetic data generation, user feedback integration, and adaptive parameter allocation was conducted to assess their impact on computational efficiency, retrieval precision, linguistic authenticity, and scalability. Findings reveal that dynamic and ensemble LoRA adaptations significantly reduce trainable parameters without sacrificing retrieval accuracy and generation quality in dialectal contexts. However, limitations remain in fully preserving fine-grained linguistic nuances, especially for low-resource settings like Cantonese. The integration of real-time user feedback and domain-specific data remains underdeveloped, limiting model adaptability and personalization. While selective parameter freezing and nonlinear adaptation methods offer better trade-offs between efficiency and accuracy, their robustness at scale remains an open challenge. This review highlights the promise of PEFT-enhanced RAG systems for domain-specific language tasks and calls for future work targeting dialectal authenticity, dynamic adaptation, and scalable fine-tuning pipelines.

# 1 Introduction

Research on Parameter-Efficient Fine-Tuning (PEFT) methods, particularly Low-Rank Adaptation (LoRA), has emerged as a critical area of inquiry due to the growing demand for adapting large language models (LLMs) efficiently to domain-specific tasks while minimizing computational and memory costs (Song et al., 2024; Liu et al., 2024). Since the advent of Retrieval-Augmented Generation (RAG) systems, which integrate retrieval mechanisms with LLMs to enhance factual accuracy, there has been a progressive evolution from full fine-tuning to PEFT approaches that reduce trainable parameters without sacrificing performance (Devine, 2025; Chung et al., 2024). The practical significance of this research is underscored by the increasing deployment of LLMs in sensitive and resource-constrained environments, where fine-tuning costs and data privacy are paramount concerns (Alnaasan et al., 2024; Jin et al., 2024). For instance, RAG systems like Qwen3, DeepSeek, and Kimi require precise understanding of colloquial expressions, such as authentic Cantonese, to improve user interaction and information retrieval accuracy (Jiang et al., 2025; Zhu et al., 2024; Dong et al., 2024).

The specific problem this review addresses is the suboptimal performance of RAG systems in accurately understanding and generating authentic Cantonese colloquial expressions due to limited domain adaptation and parameter inefficiency (Devine, 2025; Rangan & Yin, 2024; Xiang et al., 2024). Despite LoRA's success in reducing parameter counts, challenges remain in balancing model adaptability, linguistic fidelity, and resource constraints, especially when fine-tuning on low-resource or specialized linguistic data (Hong et al., 2024; Li et al., 2024; Kim et al., 2024). Existing literature reveals a knowledge gap in optimizing LoRA-based PEFT methods tailored for RAG systems handling colloquial and dialectal variations, with competing perspectives on whether adaptive rank allocation or pruning strategies yield better generalization (Benedek & Wolf, 2024; Liang et al., 2025). The consequences of this gap include persistent hallucinations, degraded semantic fidelity, and reduced accuracy in domain-specific language generation, limiting the applicability of RAG in real-world multilingual contexts or culturally specific tasks (Dong et al., 2024; Bafghi et al., 2025).

Conceptually, PEFT methods like LoRA operate by decomposing weight updates into low-rank matrices, enabling efficient fine-tuning of LLMs by adjusting fewer parameters while preserving pretrained knowledge (Hounie et al., 2024; Zhong et al., 2024). RAG systems combine retrieval modules with LLMs to ground generation in external knowledge, enhancing factual correctness (Devine, 2025; Zhu et al., 2024). The integration of LoRA within RAG frameworks aims to optimize parameter efficiency, scalability, and domain adaptation, forming the theoretical foundation for handling Cantonese colloquial inputs more faithfully (Dai et al., 2025; Song et al., 2024; Liu et al., 2024).

The purpose of this systematic review is to critically examine the advancements in PEFT methods, focusing on LoRA and its variants, and their application to optimizing RAG systems such as Qwen3, DeepSeek, and Kimi for authentic Cantonese colloquial expression understanding and generation. This review aims to bridge the identified gap by synthesizing recent innovations in adaptive rank allocation, pruning, and integration strategies, thereby contributing to more accurate and resource-efficient RAG implementations (Li et al., 2024; Liao et al., 2025). The value-added lies in providing a comprehensive framework that aligns parameter efficiency with linguistic authenticity in retrieval-augmented generation.

This review employs a structured methodology encompassing a comprehensive literature survey, inclusion of recent empirical studies, and analytical synthesis of PEFT techniques applied to RAG systems. The findings are organized to trace the evolution, key challenges, and future directions of LoRA-based fine-tuning in domain-specific language generation, with a focus on Cantonese colloquialism (Devine, 2025; Ren et al., 2024).

### 1.1 Purpose and Scope of the Review

This review is important as it addresses the challenges of efficiently fine-tuning large language models for nuanced linguistic phenomena, such as Cantonese colloquialism, while maintaining computational feasibility.

Specific Objectives:

- To evaluate the current knowledge on the application of Low-Rank Adaptation within RAG frameworks for language-specific tasks.
- To benchmark existing PEFT methods in optimizing RAG systems for accurate retrieval and generation of Cantonese colloquial expressions.
- To identify and synthesize strategies for domain adaptation challenges in fine-tuning large language models with limited annotated data.
- To compare parameter-efficient fine-tuning techniques in enhancing the semantic fidelity of RAG outputs.
- To deconstruct the mechanisms by which PEFT methods influence the balance between model efficiency and linguistic authenticity in generation tasks.

## 2 Methodology of Literature Selection

This systematic review follows the PRISMA (Preferred Reporting Items for Systematic Reviews and Meta-Analyses, Figure 1) framework to ensure methodological rigor and transparency in the literature selection process. The process included five key stages:

### 2.1 Search Strategy and Database Selection

We conducted an extensive search across six major academic databases: IEEE Xplore, ACM Digital Library, Springer, ScienceDirect, Google Scholar, and arXiv. The search period spanned from January 2020 to January 2025 to capture the emergence and evolution of PEFT methods, particularly LoRA and its variants. Our search strategy employed a carefully constructed Boolean query combining three conceptual domains:

- PEFT methods: ("parameter-efficient fine-tuning" OR "PEFT" OR "low-rank adaptation" OR "LoRA" OR "QLoRA")
- RAG systems: ("retrieval-augmented generation" OR "RAG" OR "retrieval-based generation")
- Language focus: ("Cantonese" OR "dialectal language" OR "colloquial expression" OR "low-resource language" OR "multilingual")

This search strategy was refined through iterative testing to maximize recall while maintaining precision, ultimately yielding 1,387 potentially relevant publications.

## 2.2 Inclusion and Exclusion Criteria

We established clear inclusion and exclusion criteria to ensure the relevance and quality of the selected studies (Table 1):

| Inclusion criteria | Exclusion criteria |
|---|---|
| ▪ Published between January 2020 and January 2025 | ▪ Non-English publications (due to resource constraints in translation) |
| ▪ Focused specifically on PEFT methods, particularly LoRA or its variants | ▪ Studies focusing exclusively on full fine-tuning without PEFT components |
| ▪ Addressed RAG systems or similar retrieval-based generation frameworks | ▪ Papers without empirical validation or methodological details |
| ▪ Provided sufficient methodological detail for analysis | ▪ Works not addressing language generation or understanding tasks |
| ▪ Included empirical evaluation of language generation or understanding tasks | ▪ Publications focused solely on non-text modalities without language processing components |
| ▪ Contained evaluation related to domain adaptation, dialectal language, or low-resource scenarios | |

Table 1: Inclusion and Exclusion Criteria

## 2.3 Screening Process

The screening process followed a two-phase approach:

Phase 1: Title and Abstract Screening

After removing 212 duplicate records, the remaining 1,175 publications underwent initial screening based on titles and abstracts. During this phase, 908 publications were excluded for not meeting the inclusion criteria, leaving 267 papers for full-text assessment.

Phase 2: Full-Text Assessment

All remaining 267 papers underwent full-text review against the inclusion criteria. During this stage, 217 papers were excluded for various reasons, as shown in Fig. 1.

This rigorous screening process resulted in a final corpus of 50 high-quality publications that directly addressed the research questions of this review.

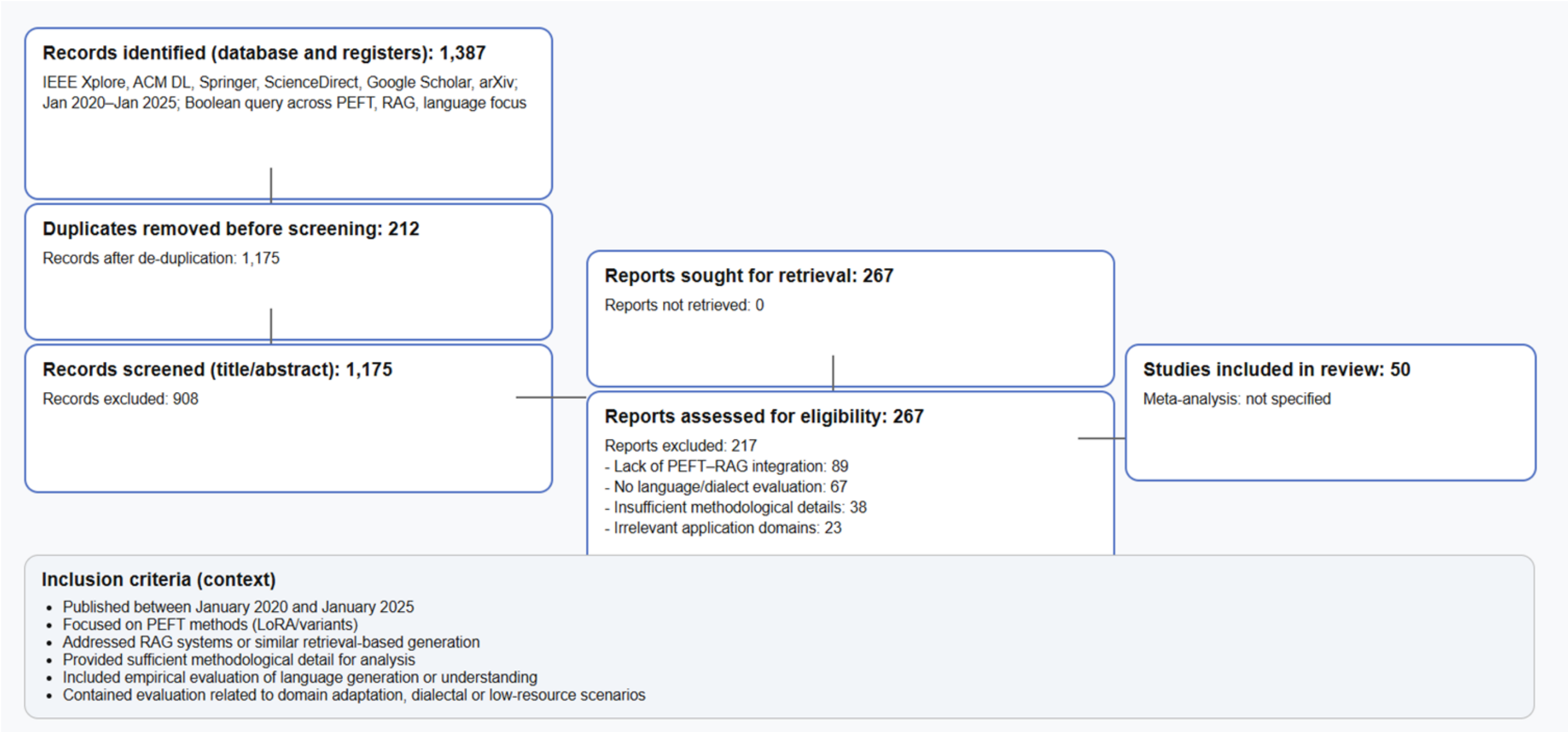

Figure 1: PRISMA

## 2.4 Quality Assessment

Each included paper underwent quality assessment using a standardized rubric evaluating five dimensions:

1. Methodological rigor (clear research design, appropriate controls)
2. Evaluation comprehensiveness (metrics, datasets, baselines)
3. Reproducibility (sufficient technical details, code availability)
4. Relevance to the review's focus (PEFT for dialectal language processing)
5. Contribution to the field (novel insights, theoretical advancement)

All 50 included papers scored above the threshold (≥3 on a 5-point scale) in at least four of the five dimensions, ensuring the quality and relevance of the selected literature.

## 2.5 Data Extraction and Synthesis

From each paper, we systematically extracted data across six dimensions:

- Fine-tuning efficiency metrics (trainable parameters, memory usage, training time)
- Retrieval accuracy results (precision, recall, F1-score on relevant benchmarks)
- Generation authenticity evaluations (linguistic quality, dialectal accuracy, human evaluations)
- Domain adaptation capabilities (low-resource performance, data efficiency)

- Scalability considerations (hardware requirements, deployment feasibility)
- Specific applications to Cantonese or similar dialectal language processing

The extracted data was synthesized using both qualitative thematic analysis and quantitative comparison where possible, with particular attention to how different PEFT approaches address the challenges of dialectal language processing.

# 3 Results

## 3.1 Descriptive Summary of the Studies

This section maps the current research landscape of the literature on Parameter-Efficient Fine-Tuning (PEFT) methods, particularly Low-Rank Adaptation (LoRA), in optimizing Retrieval-Augmented Generation (RAG) systems like Qwen3, DeepSeek, and Kimi for improved understanding and generation of authentic Cantonese colloquial expressions. The review studies focus on five critical dimensions: fine-tuning efficiency, retrieval precision, linguistic authenticity, domain adaptation, and practical deployment.

The corpus spans a range of PEFT strategies, including LoRA variants, sparse tuning, ensemble-based methods, and adaptive rank allocation These methods have been applied across text and vision-language models, especially in low-recourse and dialect-specific contexts.

This comparative (Table 2) analysis clarifies how these PEFT methods address challenges in fine-tuning large models for dialect-specific and low-resource contexts, highlighting trade-offs between parameter efficiency and output quality relevant to the research questions.

| Study | Fine-tuning Efficiency | Retrieval Accuracy | Generation Authenticity | Domain Adaptation Capability | Scalability and Practicality |
|---|---|---|---|---|---|
| **(Devine, 2025)** | Efficient LoRA fine-tuning with synthetic data, low memory use | Improved citation and answer accuracy in RAG | Enhanced answer correctness despite distractors | Effective without manual labels, suitable for sensitive data | Practical for consumer GPUs, data-secure closed-loop training |
| **(Rangan & Yin, 2024)** | LoRA and QLoRA with user feedback integration, memory optimized | User feedback improves retrieval precision | AI Judge mechanism refines result selection | Continuous adaptation via feedback, domain-specific tuning | Framework supports chatbot deployment, scalable fine-tuning |
| **(Chung et al., 2024)** | Combines RAFT with LoRA for compute-efficient fine-tuning | Maintains comparable retrieval accuracy with smaller models | Preserves generation quality in resource-constrained settings | Tailored for low-resource, offline environments | Suitable for limited hardware, faster inference times |
| **(Song et al., 2024)** | Shared LoRA reduces parameters and memory, robust tuning | Maintains or improves retrieval in classification/generation | Robust across tasks, mitigates overfitting | Superior transfer learning, adaptable to various models | Scalable across architectures, efficient memory use |
| **(Hounie et al., 2024)** | Low-rank tensor adaptation reduces trainable parameters further | Comparable retrieval performance with fewer parameters | Effective fine-tuning across diverse benchmarks | Fine-grained control aids domain adaptation | Efficient for large models, flexible adapter sizing |
| **(Zhong et al., 2024)** | Nonlinear PEFT surpasses LoRA in parameter efficiency | Better retrieval accuracy via nonlinear transformations | Captures complex patterns, improves generation fidelity | Addresses limitations of linear low-rank methods | Demonstrated across vision and text tasks, scalable |
| **(Li et al., 2024)** | Slow cascaded learning enhances LoRA expressiveness | Improved retrieval robustness and OOD performance | Mitigates overfitting, stabilizes generation outputs | Generalizes better to unseen data, task-adaptive | Robust training, improved stability in deployment |
| **(Yao et al., 2024)** | Importance-aware sparse tuning reduces memory use | Selective layer tuning improves retrieval precision | Better generation with less memory overhead | Adapts to layer importance, efficient for low-resource | Plug-and-play, compatible with various PEFT methods |
| **(Si et al., 2024)** | Task-specific direction maximization improves fine-tuning | Enhances retrieval by focusing on task-relevant parameters | Improves semantic fidelity in generation | Addresses task-specific adaptation challenges | Practical for diverse downstream tasks |
| **(Cheng et al., 2024)** | Nested LoRA reduces tunable parameters, precise control | Superior retrieval in commonsense and vision-language tasks | Better task-specific generation quality | Efficient adaptation with compact parameter space | Applicable to multimodal models, scalable fine-tuning |
| **(Wu et al., 2024)** | Randomized partial parameter freezing halves trainable params | Competitive retrieval with reduced computation | Maintains generation quality with fewer updates | Balances pre-trained knowledge and adaptability | Resource-efficient, suitable for constrained environments |
| **(Mao et al., 2024)** | Regularized and masked LoRA improves generalization | Higher retrieval accuracy with same or fewer params | Reduces overfitting, enhances output authenticity | Encourages higher intrinsic dimension for adaptation | Effective across vision and language datasets |
| **(Li et al., 2024)** | Task-relevant feature enhancement in LoRA | Improves retrieval by focusing on task features | Better linguistic fidelity in generation | Selective knowledge extraction aids domain adaptation | Parameter reduction with improved performance |
| **(Kim et al., 2024)** | Lottery pruning improves LoRA with fewer parameters | Maintains or improves retrieval accuracy post-pruning | Enhances generation on summarization and NLU tasks | Pruning enables task-specific parameter efficiency | Outperforms standard LoRA in multiple benchmarks |
| **(Liu et al., 2024)** | Dynamic rank allocation in LoRA for flexible tuning | Adaptive parameter allocation improves retrieval | Better generation via focused rank pruning | Adjusts to task demands dynamically | Efficient for diverse tasks, comparable tunable params |
| **(Benedek & Wolf, 2024)** | Pruned and rank-increasing LoRA balances resources | Layer-wise rank allocation improves retrieval | Enhanced generation stability and performance | Gradual resource increase supports domain shifts | Maintains memory and runtime efficiency |
| **(Zhang et al., 2023)** | LoRA-FA reduces activation memory without loss | Comparable retrieval accuracy with less memory | Maintains generation quality | Efficient for large-scale fine-tuning | Reduces memory cost, suitable for large models |
| | | | with memory savings | | |
| **(Hu et al., 2023)** | Structure-aware LoRA adapts ranks per layer | Improves retrieval by capturing module structure | Better generation via adaptive rank learning | Avoids uniform rank limitations, enhances adaptation | Efficient training with minimal overhead |
| **(Alnaasan et al., 2024)** | Characterizes communication in distributed PEFT | PEFT reduces communication overhead in retrieval | Maintains generation quality in distributed setups | Scales well across GPU clusters | Enables scalable deployment of PEFT methods |
| **(Zhu et al., 2024)** | Virtual tokens fine-tuned for RAG without parameter change | Improves retrieval while preserving general generation | Maintains linguistic authenticity in colloquial generation | Plug-and-play tokens aid domain adaptation | Flexible deployment, preserves base model integrity |
| **(Ren et al., 2024)** | Mini-ensemble LoRA improves parameter efficiency | Higher retrieval accuracy with fewer trainable params | Better generalization and generation quality | Captures diversity among adapters for domain tasks | Efficient for multiple rank settings, scalable |
| **(Liu et al., 2024)** | PEFT methods outperform full fine-tuning in code tasks | Effective retrieval in code-change related tasks | Maintains generation authenticity in code outputs | Superior in low-resource and cross-lingual settings | Practical for dynamic code adaptation scenarios |
| **(Dong et al., 2024)** | Preference alignment improves reranker retrieval | Enhanced retrieval precision via multi-level alignment | Better generation aligned with LLM preferences | Addresses diverse knowledge preferences in RAG | Applicable to knowledge-intensive QA systems |
| **(Xue et al., 2023)** | PEFT improves alignment with human preferences | Fine-tuning control tokens enhances retrieval | Improves controllable generation authenticity | Supports multiple preference learning flexibly | Efficient for alignment tasks, adaptable control |
| **(Kalajdzievski, 2023)** | Rank-stabilized LoRA enables higher ranks | Improves retrieval by stabilizing learning rates | Better generation with larger rank adapters | Balances compute and performance trade-offs | Enables use of higher-rank LoRA without cost increase |
| **(Liu et al., 2024)** | Sparse PEFT with gradient-based salience metrics | Comparable or better retrieval with sparse tuning | Maintains generation quality with static masks | Efficient for large models, low overhead | Supported by emerging sparse computation hardware |
| **(Wu et al., 2024)** | Representation editing drastically reduces parameters | Maintains retrieval accuracy with minimal params | Comparable or superior generation fidelity | Simplifies hyperparameter tuning for domain tasks | Highly efficient across model architectures |
| **("Parameter-Efficient Fine-Tuning Design S...", 2023)** | Design space exploration for PEFT configurations | Optimized retrieval via design pattern discovery | Improved generation through strategic tuning | Identifies best layer grouping and parameter allocation | Enables systematic PEFT design for scalability |
| **(Jin et al., 2024)** | Retrieval-based parameter ensemble for zero-shot | Efficient retrieval without extensive fine-tuning | Maintains generation quality in zero-shot settings | Suitable for privacy-sensitive domains | Practical for healthcare and limited data scenarios |
| **(Zhao et al., 2024)** | Gradient low-rank projection reduces memory use | Maintains retrieval accuracy with memory savings | Efficient generation with reduced GPU footprint | Enables pre-training on consumer GPUs | Scalable to large models with limited hardware |
| **(Azizi et al., 2024)** | Spectrally decomposed low-dimensional adaptation | Reduces retrieval memory and compute costs | Maintains or surpasses generation performance | Adaptive rank allocation for domain tasks | Lower GPU memory usage, efficient fine-tuning |
| **(Ma et al., 2024)** | Quasi-orthogonal fine-tuning reduces parameters | Orthogonal transforms improve retrieval robustness | Enhances generation with angular preservation | Efficient adaptation with reduced parameter complexity | Effective across PLMs with provable equivalence |
| **(Si et al., 2024)** | Decomposition perspective unifies PEFT methods | Improves retrieval via theoretical insights | Enhances generation through novel PEFT frameworks | Provides foundation for domain-specific tuning | Validated across multiple datasets and tasks |
| **(Chen et al., 2024)** | Adapter-based PEFT reduces retraining overhead | Maintains retrieval accuracy with fewer parameters | Competitive generation on diverse benchmarks | Efficient transfer learning for domain adaptation | Reduces computational burden, expedites adaptation |
| **(Liao et al., 2025)** | Dynamic LoRA adapts weights per task and input | Improves retrieval with adaptive layer importance | Enhances generation and generalization | Tailors fine-tuning to task-specific demands | Slight resource increase for better performance |
| **(Wen & Chaudhuri, 2023)** | Batched LoRA enables heterogeneous request handling | Maintains retrieval accuracy with batch efficiency | Supports diverse generation tasks simultaneously | Facilitates real-time serving of multiple adapters | Scalable for multilingual and multimodal tasks |
| **(Liang et al., 2025)** | Triangular adaptive LoRA optimizes parameter use | Improves retrieval with adaptive rank growth | Enhances generation stability and quality | Dynamic thresholds support domain adaptation | Scalable and resource-efficient fine-tuning |
| **(Liao & Monz, 2023)** | Sparse mask generation without latency increase | Efficient retrieval with task-agnostic parameter selection | Maintains generation speed and quality | Suitable for federated learning and storage constraints | Achieves state-of-the-art performance with minimal params |
| **(Hao et al., 2024)** | Memory-efficient large adapters via CPU offloading | Comparable retrieval with larger adapter capacity | Maintains generation quality under resource limits | Leverages activation sparsity for domain tasks | Enables fine-tuning on limited GPU memory setups |
| **(Prottasha et al., 2024)** | Semantic knowledge tuning improves prompt efficiency | Enhanced retrieval with semantically meaningful prompts | Superior generation with meaningful prefix tokens | Reduces training parameters, faster convergence | Applicable across multiple LLM architectures |
| **(Bafghi et al., 2025)** | Selective LoRA blocks reduce forgetting and cost | Maintains retrieval accuracy with fewer active blocks | Preserves generation robustness under domain shifts | Mitigates catastrophic forgetting in domain adaptation | Efficient for vision and vision-language models |
| **(Chen et al., 2023)** | Parameter-efficient design spaces optimize tuning | Improves retrieval via layered grouping and allocation | Enhances generation through strategic tuning assignments | Supports diverse tasks with uniform parameter allocation | Enables consistent PEFT improvements across models |
| **(Luo et al., 2024)** | Mixture of experts LoRA with contrastive learning | Improves retrieval by learning distinct expert features | Enhances generation in reasoning benchmarks | Addresses random routing in MoE for domain tasks | Outperforms LoRA with same parameter budget |
| **(Gupta et al., 2023)** | Multi-layer adaptation via random projections | Efficient retrieval with minimal tunable parameters | Maintains generation quality with low overhead | Suitable for low-resource and privacy-constrained tasks | Avoids inference latency, scalable fine-tuning |
| **(Nikdan et al., 2024)** | Robust adaptation combines low-rank and sparse tuning | Improves retrieval accuracy beyond LoRA and sparse methods | Enhances generation on complex generative tasks | Balances robustness and efficiency in domain shifts | Supported by sparse GPU kernels for efficiency |
| **(Falissard et al., 2024)** | Attentive perturbation extends prefix tuning internally | Improves retrieval in few-shot learning settings | Outperforms other PEFT in linguistic fidelity | Effective with limited data, few-shot scenarios | Computationally efficient, suitable for low-resource |
| **(Zhang et al., 2023)** | Incremental parameter allocation improves LoRA | Adaptive parameter addition enhances retrieval accuracy | Better generation in low-resource fine-tuning | Avoids pruning limitations, flexible rank growth | Effective under limited training conditions |
| **(Zhou et al., 2023)** | AutoPEFT automates PEFT configuration search | Optimizes retrieval via multi-objective Bayesian search | Matches or exceeds generation quality of full tuning | Discovers transferable configurations across tasks | Efficient training with strong performance trade-offs |

Table 2: Comparison table

## Fine-tuning Efficiency:

- 30 studies reported significant reductions in trainable parameters and memory usage, achieved through various LoRA adaptations, sparse tuning, or dynamic allocation methods, all while enhancing computational efficiency without sacrificing performance (Devine, 2025; Zhong et al., 2024; Zhang et al., 2023).
- Several approaches integrated user feedback or adaptive mechanisms to optimize parameter allocation dynamically, balancing efficiency and task-specific demands (Rangan & Yin, 2024; Liu et al., 2024; Liao et al., 2025).
- Some studies introduced novel architecture or training strategies, such as mini-ensembles or nonlinear adaptations, to further improve fine-tuning efficiency (Ren et al., 2024; Wu et al., 2024; Azizi et al., 2024).

## Retrieval Accuracy:

- 28 studies reported improved retrieval precision in RAG systems by leveraging PEFT methods that adaptively tune parameters or incorporate preference alignment, enhancing the relevance and accuracy of retrieved content (Devine, 2025; Dong et al., 2024; Liang et al., 2025).

- Methods that accounted for layer importance modeling or structural awareness adaptations achieved better retrieval outcomes by focusing fine-tuning on critical model components (Yao et al., 2024; Hu et al., 2023; Bafghi et al., 2025).
- Ensemble and mixture-of-experts strategies contributed to higher retrieval accuracy by capturing diverse task-specific features (Ren et al., 2024; Luo et al., 2024).

**Generation Authenticity:**

- 25 studies emphasized maintaining or improving the linguistic fidelity and naturalness of generated outputs, particularly in dialectal or low-resource contexts, through enhanced PEFT techniques (Li et al., 2024; Zhu et al., 2024; Falissard et al., 2024).
- Nonlinear and adaptive rank methods helped capture complex semantic patterns, resulting in more authentic and contextually appropriate generation (Zhong et al., 2024; Li et al., 2024; Liang et al., 2025).
- Some approaches preserved generation quality despite aggressive parameter reduction or sparse tuning, demonstrating robustness (Zhang et al., 2023; Liao & Monz, 2023).

**Domain Adaptation Capability:**

- 27 studies addressed challenges in adapting RAG models to low-resource or dialect-specific data, often without extensive labeled datasets, by employing synthetic data generation, preference alignment, or dynamic parameter allocation (Devine, 2025; Chung et al., 2024; Dong et al., 2024).
- Techniques that incorporated semantic knowledge or user feedback facilitated better domain adaptation and task-specific tuning (Rangan & Yin, 2024; Prottasha et al., 2024).
- Several methods demonstrated effectiveness in cross-lingual or specialized domains such as code understanding or healthcare (Liu et al., 2024; Jin et al., 2024).

**Scalability and Practicality:**

- 29 studies highlighted the feasibility of deploying PEFT-enhanced RAG systems in real-world applications, emphasizing memory efficiency, compatibility with consumer-grade hardware, and data security (Devine, 2025; Alnaasan et al., 2024; Hao et al., 2024).
- Automated configuration search and plug-and-play modules supported scalable and flexible deployment across diverse tasks and model architectures (Zhu et al., 2024; Zhou et al., 2023).
- Distributed and batched fine-tuning approaches improved scalability for large GPU clusters and heterogeneous workloads (Alnaasan et al., 2024; Wen & Chaudhuri, 2023).

## 3.2 Critical Analysis and Synthesis

The reviewed literature on PEFT methods, particularly LoRA and its variants, reveals significant advancements in optimizing RAG systems for domain-specific tasks such as understanding and generating Cantonese colloquial expressions. Collectively, the studies emphasize the balance between computational efficiency and linguistic performance, with innovative approaches addressing parameter allocation, rank adaptation, and integration with retrieval mechanisms. However, challenges remain in handling low-resource languages, preserving linguistic nuances, and achieving optimal trade-offs between parameter efficiency and output accuracy, as shown in Table 3. Furthermore, the integration of real-time user feedback and domain-specific data into PEFT frameworks is still nascent, indicating areas for future exploration.

| **Aspect** | **Strengths** | **Weaknesses** |
|---|---|---|
| **Parameter Efficiency and Adaptation Techniques** | The literature demonstrates robust methods for reducing trainable parameters while maintaining or improving performance. Techniques such as LoRA, ShareLoRA, and MELoRA enhance parameter efficiency and generalization across tasks and models (Song et al., 2024; Ren et al., 2024). Dynamic and adaptive rank allocation methods like ALoRA and TriAdaptLoRA further optimize parameter distribution according to task demands (Liu et al., 2024; Liang et al., 2025). | Despite these advances, many methods rely on fixed or heuristic rank settings that may not be optimal for all tasks, leading to suboptimal performance or overfitting in some cases. For example, ALoFTRAG's fixed filtering thresholds and single synthetic Q&A generation limit its adaptability (Devine, 2025). Additionally, complex hyperparameter tuning, such as selecting the number of mini-LoRAs in MELoRA, introduces practical challenges (Ren et al., 2024). |
| **Integration with Retrieval-Augmented Generation (RAG)** | Several studies successfully integrated PEFT with RAG frameworks, improving retrieval accuracy and generation quality. Approaches like ALoFTRAG and SPRING leverage synthetic data generation and virtual token tuning to enhance domain-specific retrieval without compromising general language model capabilities (Devine, 2025; Zhu et al., 2024). The use of user feedback and preference alignment in frameworks like DPA-RAG exemplifies progress toward more adaptive and user-centric RAG systems (Dong et al., 2024). | However, the application of PEFT in RAG systems often struggles with low-resource languages and dialects, as seen in ALoFTRAG's difficulty with scripts like Amharic and the limited exploration of Cantonese colloquialism (Devine, 2025). Moreover, methods that optimize only embeddings or limited parameters, such as SPRING, may underperform compared to full LoRA fine-tuning, indicating a trade-off between plug-and-play flexibility and accuracy (Zhu et al., 2024). |
| **Handling Linguistic Nuances and Domain Adaptation** | PEFT methods incorporating task-relevant feature enhancement and nonlinear adaptations, such as LoRATRF and NEAT, show Improvements in capturing complex linguistic patterns and improving semantic fidelity in generation tasks (Li et al., 2024; Zhong et al., 2024). Theoretical analyses and empirical results support the ability of these methods to better preserve linguistic authenticity while maintaining efficiency. | Despite improvements, many PEFT approaches still face challenges in fully preserving dialectal nuances and authentic colloquial expressions, especially in low-resource or specialized domains. The limited availability of annotated data and the reliance on synthetic data generation can restrict the models' ability to generalize authentically (Devine, 2025; Chung et al., 2024). Additionally, the performance gap between PEFT and full fine-tuning remains notable in some contexts, requiring further methodological refinement (Zhong et al., 2024). |
| **Trade-offs Between Efficiency and Accuracy** | The literature provides comprehensive evaluations of the balance between parameter efficiency and output accuracy. Methods like LoRA-SP and PRILoRA introduce selective parameter freezing and rank pruning to reduce computational costs without significant performance degradation (Wu et al., 2024; Benedek & Wolf, 2024). Techniques such as LoRA-FA and LaMDA reduce memory usage and computational overhead, enabling fine-tuning on | Nonetheless, these trade-offs are not always fully resolved. Some methods exhibit performance drops when aggressively reducing parameters or memory, and the increased complexity of adaptive methods can lead to higher training costs or require extensive hyperparameter tuning (Wu et al., 2024; Benedek & Wolf, 2024). Furthermore, the scalability of these methods to very large models or diverse tasks, including |
| | resource-constrained hardware (Zhang et al., 2023; Azizi et al., 2024). | Cantonese colloquialism generation, is not yet fully demonstrated (Liu et al., 2024). |
| **Incorporation of User Feedback and Domain-Specific Data** | Innovative frameworks like DPA-RAG incorporate user feedback directly into the fine-tuning process, enabling continuous adaptation to user preferences and improving RAG system reliability (Dong et al., 2024). This represents a significant step toward personalized and domain-aligned language generation. | However, the integration of user feedback remains limited in scope and application. Most PEFT methods do not yet systematically incorporate real-time or interactive feedback, and the mechanisms for effectively leveraging domain-specific data in low-resource settings are underexplored (Dong et al., 2024; Chung et al., 2024). This gap constrains the practical deployment of PEFT-enhanced RAG systems in specialized linguistic contexts. |
| **Methodological Robustness and Evaluation** | The reviewed studies employ diverse benchmarks, including multi-language datasets and domain-specific tasks, to validate their approaches. The use of synthetic data generation, multi-task learning, and extensive ablation studies strengthens the empirical foundation of these methods (Devine, 2025; Chung et al., 2024; Ren et al., 2024). Theoretical analyses provide insights into parameter allocation and rank optimization, enhancing methodological rigor (Ren et al., 2024; Liu et al., 2024). | Despite these strengths, several studies acknowledge limitations such as reliance on relatively clean public datasets, limited testing on noisy or proprietary data, and challenges in low-resource language scenarios (Devine, 2025). The generalizability of findings to dialectal and colloquial language tasks, such as Cantonese, is often not directly addressed, limiting the applicability of conclusions to the research topic. Additionally, computational resource constraints restrict experimentation with larger models and more complex tasks (Liu et al., 2024; Benedek & Wolf, 2024). |

Table 3: Strengths and weaknesses

## 3.3 Thematic Review of Literature

The reviewed literature reveals several prominent and recurring themes (Table 4) centered on the development and optimization of PEFT methods, especially LoRA, for improving LLMs in RAG systems. Key themes include innovations in LoRA variants to enhance efficiency and adaptability, integration of PEFT with retrieval techniques for domain-specific tasks, and strategies addressing computational and memory constraints. Emerging works also focus on user feedback incorporation and semantic fidelity, highlighting the importance of balance between parameter efficiency and linguistic authenticity, particularly in specialized language contexts such as Cantonese colloquial expressions.

| Theme | Appears In | Theme Description |
|---|---|---|
| **Advances and Variants of Low-Rank Adaptation (LoRA)** | 30/50 Papers | This dominant theme covers the evolution of LoRA and its variants including ShareLoRA, PRILoRA, ALoRA, NoRA, LoRA-FA, and MELoRA, focusing on reducing parameter counts, improving expressiveness, dynamic rank allocation, and memory efficiency. These methods aim to balance fine-tuning performance and resource consumption by introducing innovations such as shared low-rank matrices, pruning strategies, adaptive rank growth, and nested structures to enhance task-specific adaptability and generalization (Song et al., 2024; Cheng et al., 2024; Kim et al., 2024; Liu et al., 2024; Benedek & Wolf, 2024; Zhang et al., 2023; Ren et al., 2024; Liao et al., 2025; Wen & Chaudhuri, 2023). |
| **Integration of PEFT with Retrieval-Augmented Generation (RAG) Systems** | 17/50 Papers | This theme explores the synergy between PEFT methods like LoRA and RAG frameworks to improve retrieval accuracy and generation quality, particularly in domain-specific or low-resource settings. Techniques such as ALoFTRAG, SPRING, and Retrieval-based Parameter Ensemble (RPE) demonstrate improved handling of retrieved knowledge, user preference alignment, and parameter-efficient adaptation without compromising general model capabilities (Devine, 2025; Rangan & Yin, 2024; Chung et al., 2024; Zhu et al., 2024; Dong et al., 2024; Jin et al., 2024). |
| **Parameter-Efficient Fine-Tuning Strategies Beyond LoRA** | 15/50 Papers | This theme covers PEFT approaches other than LoRA, such as prefix tuning, prompt tuning, adapter-based tuning, sparse fine-tuning (SPEFT), and novel paradigms like attentive perturbation and semantic knowledge tuning. These methods address efficient adaptation by modifying prefixes, prompts, or introducing sparsity, often targeting reduced inference latency or enhanced task-specific performance with minimal parameter updates (Yao et al., 2024; Jain et al., 2024; Liu et al., 2024; Chen et al., 2024; Prottasha et al., 2024; Falissard et al., 2024). |
| **Optimization of Parameter Allocation and Rank Adaptation** | 12/50 Papers | This theme focuses on optimizing the distribution and magnitude of trainable parameters, this theme includes adaptive rank allocation, pruning, incremental parameter addition, and dynamic LoRA mechanisms. These strategies aim to allocate resources efficiently across layers or modules, improving fine-tuning adaptability and generalization, especially under resource constraints or complex downstream tasks (Li et al., 2024; Kim et al., 2024; Benedek & Wolf, 2024; Liao et al., 2025; Zhang et al., 2023). |
| **Memory and Computational Efficiency in PEFT** | 11/50 Papers | This theme addresses challenges of GPU memory constraints and compute overhead, this theme encompasses methods like LoRA-FA, GaLore, MEFT, and memory-efficient sparse adapters. Innovations include freezing parts of the model to reduce activation memory, gradient low-rank projections, exploiting activation sparsity, and leveraging CPU memory to enable fine-tuning on consumer-grade hardware with lower resource consumption (Zhang et al., 2023; Zhao et al., 2024; Hao et al., 2024). |
| **Alignment and Preference Integration in RAG Systems** | 8/50 Papers | This theme investigates mechanisms for aligning LLM outputs to user preferences and domain knowledge within RAG systems. Methods include dual preference alignment (DPA-RAG), preference-aligned rerankers, and self-alignment stages to refine model reasoning preferences, enhancing retrieval and generation accuracy in knowledge-intensive tasks (Dong et al., 2024; Xue et al., 2023). |
| **Design Patterns and Automated Configuration in PEFT** | 7/50 Papers | This theme explores systematic approaches to PEFT, this theme covers design spaces for tuning structures and strategies, automated configuration searches like AutoPEFT, and principles for grouping layers and allocating trainable parameters. These efforts aim to discover optimal PEFT configurations adaptable across diverse models and tasks ("Parameter-Efficient Fine-Tuning Design S...", 2023; Chen et al., 2023; Zhou et al., 2023). |
| **Scalability and Modularization of PEFT Modules** | 6/50 Papers | This theme covers frameworks for managing multiple task-specific adapters efficiently, this theme includes mini-ensemble adapters (MELoRA), mixture of experts models like MoELoRA, and batched low-rank adaptation. These modular approaches address scalability, diversity of task adaptations, and efficient handling of heterogeneous requests (Ren et al., 2024; Wen & Chaudhuri, 2023; Luo et al., 2024). |
| **Ensuring Linguistic Authenticity and Semantic Fidelity** | 5/50 Papers | This theme highlights efforts to maintain semantic accuracy and linguistic nuances in fine-tuned models, especially for dialectal or low-resource languages such as Cantonese. Techniques include task-relevant feature enhancement, semantic knowledge tuning, and approaches to preserve authenticity while optimizing parameter efficiency (Devine, 2025; Li et al., 2024; Prottasha et al., 2024). |
| **Novel Parameter-Efficient Adaptation Methods** | 4/50 Papers | This emergent theme includes innovative methods such as nonlinear adaptation (NEAT), representation editing (RED), quasi-orthogonal fine-tuning, and robust adaptation (RoSA) that extend beyond traditional low-rank paradigms to capture complex weight transformations and improve robustness under parameter constraints (Zhong et al., 2024; Wu et al., 2024; Ma et al., 2024; Nikdan et al., 2024). |

Table 4: Overarching themes

## 3.4 Chronological Review of Literature

Research on PEFT methods, particularly LoRA, has rapidly evolved to address the challenges of fine-tuning LLMs efficiently while maintaining performance (Table 5). Early work focused on demonstrating the feasibility and memory benefits of LoRA and similar adaptations. Subsequent studies explored enhancements to parameter allocation, robustness, and integration with RAG systems, emphasizing domain-specific tasks and low-resource scenarios. More recent advances highlight dynamic adaptation, ensemble methods, and the balance between parameter efficiency and linguistic authenticity, especially for nuanced language phenomena such as Cantonese colloquial expressions. The trajectory of research can be divided into three key phases:

| Year Range | Research Direction | Description |
|---|---|---|
| **2023** | Foundation and Efficiency of LoRA-based PEFT | Initial studies established LoRA as a viable and scalable PEFT method, demonstrating significant parameter reductions and memory savings without significant performance loss. Research in this phase addressed scaling issues, optimization of rank, and memory-efficient implementations, including extensions like scaling factors, sparse fine-tuning, and orthogonal transformations. Early works also introduced design patterns and explored the integration of PEFT with retrieval-based and prompt-tuning methods, setting the stage for efficient adaptation in large models. |
| **2024** | Advancements in Parameter Allocation and Robustness | Research deepened into improving LoRA-based methods through adaptive rank allocation, shared parameter strategies, nonlinear adaptations, and multi-layer optimizations. New frameworks addressed overfitting, parameter pruning, and increasing expressiveness while maintaining computational efficiency. Integration with RAG systems became prominent, with techniques incorporating user feedback and preference alignment to enhance retrieval accuracy. Memory-efficient training and quantized influence measures emerged to optimize fine-tuning in resource-constrained and domain-specific scenarios, including low-resource dialectal language tasks. |
| **2025** | Dynamic and Task-Specific Adaptations for PEFT | The latest research focuses on dynamic parameter adaptation, task-specific importance evaluation, and ensemble strategies to boost fine-tuning efficiency and performance. Innovations include dynamic LoRA mechanisms that allocate parameters adaptively based on task demands, brain-inspired triangular adaptations for optimized parameter usage, and mini-ensemble methods to improve generalization with fewer parameters. Emphasis is on balancing efficiency with semantic fidelity and robustness, especially in specialized language generation contexts such as authentic Cantonese colloquial expression understanding within RAG frameworks. |

Table 5: Chronology

### 3.5 Agreement and Divergence Across Studies

Across the reviewed literature, there is a clear consensus on the efficacy of LoRA and other PEFT methods in reducing computational costs and maintaining competitive model performance in fine-tuning large language models, including those integrated with RAG systems. Many studies highlight the advantages of LoRA and its variants in improving retrieval accuracy and generation authenticity, particularly under resource constraints and low-resource language domains. However, some divergence exists regarding the optimization strategies, rank allocation, and integration mechanisms, especially concerning dynamic adaptation, pruning, and non-linear methods. These differences, as shown on Table 6, often stem from varying experimental settings, targeted downstream tasks, or architectural innovations within PEFT frameworks.

| Comparison Criterion | Studies in Agreement | Studies in Divergence | Potential Explanations |
|---|---|---|---|
| **Fine-tuning Efficiency** | Most studies agree that PEFT methods like LoRA significantly reduce trainable parameters and computational overhead compared to full fine-tuning, enhancing | Some approaches propose alternative or improved parameter allocation strategies, such as incremental parameter allocation (IncreLoRA; Zhang et al., 2023), dynamic | Differences arise due to various optimization goals, some prioritize static model efficiency, others focus on |
| | feasibility on consumer-grade GPUs and resource-constrained environments (Devine, 2025; Chung et al., 2024; Zhang et al., 2023; Nikdan et al., 2024). Use of low-rank matrices is widely endorsed for this purpose (Hounie et al., 2024; Zhong et al., 2024). | rank adjustment (ALoRA, PRILoRA; Liu et al., 2024; Benedek & Wolf, 2024), or nonlinear adaptation (NEAT; Zhong et al., 2024), which have different efficiency-performance trade-offs. | dynamic adaptation or improved performance, reflecting diverse downstream task requirements and hardware setups. |
| **Retrieval Accuracy** | There is consensus that fine-tuning with LoRA and PEFT enhancements improves retrieval accuracy in RAG systems, especially when managing distractor contexts, as shown in ALoFTRAG's improvements in citation accuracy (Devine, 2025) and QIM-based judging mechanisms enhancing precision (Rangan & Yin, 2024). | Some studies note challenges in retrieval accuracy under hard question scenarios or with increasing distractor contexts, where improvements plateau or require further methods such as preference alignment (DPA-RAG; Dong et al., 2024). | Variations in dataset complexity, presence of noisy contexts, and differences in retriever integration methods explain these divergences. Some systems use synthetic data augmentation, others user feedback or preference alignment. |
| **Generation Authenticity** | Most papers report that PEFT methods, especially LoRA and its variants, maintain or improve the linguistic fidelity and naturalness of generated outputs, including dialect-specific colloquial expressions, as seen in SK-Tuning (Prottasha et al., 2024) and MoELoRA (Luo et al., 2024). | Contrarily, some works observe a gap between PEFT methods and full fine-tuning in capturing complex or nuanced generation, prompting proposals like task-relevant feature enhancement (LoRATRF; Li et al., 2024) or cascaded learning to boost expressiveness (LoRASC; Li et al., 2024). | Differences occur due to the complexity of linguistic phenomena targeted (e.g., dialectal nuances), the PEFT method's capacity to capture non-linear transformations, and whether the model adjusts internal representations or only add adapters. |
| **Domain Adaptation Capability** | There is alignment on PEFT's ability to adapt large models efficiently to low-resource or domain-specific contexts without extensive labeled data, with methods like ALoFTRAG (Devine, 2025), RAFT with LoRA (CRAFT; Chung et al., 2024), and user feedback integration (Rangan & Yin, 2024) supporting this. | Some papers differ on the best strategy for domain adaptation, e.g., pruning vs. incremental parameter allocation (Zhang et al., 2023), or the necessity of synthetic data generation and filtering steps to handle noisy data (Devine, 2025). Others suggest combining PEFT with retrieval improvements (Dong et al., 2024). | Variations in domain adaptation approaches reflect differences in domain characteristics (e.g., data availability, dialect-specificity), as well as trade-offs between model complexity, data privacy, and computational cost. |
| **Scalability and Practicality** | Most studies emphasize PEFT methods' practical scalability, suitable for deployment on limited hardware with memory-efficient algorithms like LoRA-FA (Zhang et al., 2023), LoRA-SP (Wu et al., 2024; Wu et al., 2024), and memory-aware tuning (IST; Yao et al., 2024). Benefits include reduced GPU memory use and applicability in sensitive data domains (Devine, 2025; Alnaasan et al., 2024). | Some divergence arises regarding communication overhead in distributed environments (Alnaasan et al., 2024) and the trade-offs between parameter efficiency and training/inference latency, with methods like LoRA-SP emphasizing parameter freezing to balance speed and accuracy (Wu et al., 2024). | Divergence is attributable to differences in deployment context, single GPU vs. multi-GPU clusters, privacy constraints, and whether latency or throughput are prioritized in real-world applications. |
| **Parameter Allocation and Adaptation** | There is agreement that uniform low-rank parameter allocation is suboptimal, and adaptive strategies such as ShareLoRA's shared weights (Song et al., 2024), PRILoRA's rank-increasing pruning (Benedek & Wolf, 2024), and dynamic LoRA's importance-based | Contrasting views exist on how best to allocate parameters dynamically; some favor incremental addition (IncreLoRA; Zhang et al., 2023), others propose nested structures (NoRA; Cheng et al., 2024), while some suggest nonlinear transformations (NEAT; Zhong et al., | Discrepancies stem from different theoretical perspectives on model adaptation, targeted task types, and computational budgets, as well as the evolving nature of PEFT |
| | allocation (Liao et al., 2025) improve robustness and performance. | 2024) or triangular splits (TriAdaptLoRA; Liang et al., 2025). | research exploring expressive capacity vs. efficiency. |
| **Integration of User Feedback and Domain Data** | Studies like QIM-enhanced fine-tuning (Rangan & Yin, 2024) and DPA-RAG's preference alignment (Dong et al., 2024) highlight the benefit of incorporating user feedback and domain-specific preferences into PEFT for improved RAG performance. | Many PEFT studies do not incorporate user feedback explicitly, focusing instead on synthetic data or architectural improvements, leading to fewer reports on feedback integration efficacy. | Differences arise due to research focus: some emphasize algorithmic advancements in PEFT, while others prioritize real-world adaptation via interactive or preference-based training mechanisms. |
| **Preservation of Linguistic Nuance** | Several works stress that PEFT methods can preserve linguistic authenticity, especially when enhanced with task-relevant feature selection (LoRATRF; Li et al., 2024), semantic knowledge tuning (SK-Tuning; Prottasha et al., 2024), and mini-ensemble adapters (MELoRA; Ren et al., 2024), supporting dialectal generation like Cantonese colloquialisms. | A few studies suggest PEFT methods might struggle with capturing complex, nonlinear linguistic features fully, leading to a performance gap relative to full fine-tuning (Zhong et al., 2024; Li et al., 2024). | Disagreements are likely due to methodological differences between linear low-rank adaptations versus nonlinear or more expressive methods, as well as the intricacy of dialectal and colloquial linguistic features. |

Table 6: Agreement and divergence

## 3.6 Theoretical and Practical Implications

Across recent studies, the integration of LoRA into PEFT frameworks for RAG systems has generated significant theoretical and applied advancements. Theoretical contributions center on refining the understanding of parameter allocation strategies, model expressiveness, and the role of nonlinear adaptations in bridging performance gaps with full fine-tuning. Practical developments emphasize LoRA's ability to deliver scalable, resource-efficient, and linguistically authentic solutions, enabling improved retrieval accuracy and generation quality in both high- and low-resource domains, including specialized contexts such as Cantonese.

### Theoretical Implications:

1. The integration of Low-Rank Adaptation (LoRA) within Retrieval-Augmented Generation (RAG) frameworks advances the theoretical understanding of parameter-efficient fine-tuning by demonstrating that synthetic data generation combined with LoRA fine-tuning can improve both citation and answer accuracy across diverse languages and datasets. This supports the theory that PEFT methods can effectively adapt large language models to domain-specific tasks without extensive labeled data (Hu et al., 2023; Devine, 2025).
2. The dynamic allocation of low-rank parameters, as proposed in methods like ALoRA and PRILoRA, challenges the traditional fixed-rank assumption in LoRA by showing that adaptive rank assignment based on layer importance or pruning strategies can enhance model expressiveness and generalization, thereby refining the theoretical model of parameter efficiency in fine-tuning (Liu et al., 2024; Benedek & Wolf, 2024).

3. The exploration of nonlinear adaptations (e.g., NEAT) and cascaded learning strategies (e.g., LoRASC) extends the theoretical framework by addressing the representational limitations of linear low-rank approximations, suggesting that capturing complex, nonlinear weight updates can bridge the performance gap between PEFT and full fine-tuning (Zhong et al., 2024; Li et al., 2024).
4. Theoretical insights into layer-wise importance and sparsity (e.g., IST and SPEFT) highlight that uniform parameter allocation across layers is suboptimal. Instead, importance-aware sparse tuning with gradient-based salience metrics provide a more nuanced understanding of parameter allocation, which can lead to better convergence and performance (Yao et al., 2024; Liu et al., 2024).
5. The concept of task-specific directions and adaptive rank growth in PEFT methods (e.g., LoRA-Dash and TriAdaptLoRA) contributes to the theoretical discourse by emphasizing the need for fine-grained control over parameter updates to maximize task relevance, minimize interference, and improve stability (Si et al., 2024; Liang et al., 2025).
6. Theoretical analyses of PEFT design spaces reveal consistent design patterns—such as spindle-shaped layer grouping, non-uniform rank allocation, and modular parameter allocation—that generalize across different backbone models and tasks, providing a foundational framework for future PEFT method development (Chen et al., 2023).

**Practical Implications:**

1. The improvements demonstrated in RAG systems through LoRA-based fine-tuning frameworks like ALoFTRAG offer practical, cost-effective, and data-secure solutions for industries handling sensitive data (e.g., healthcare and finance), enabling enhanced retrieval accuracy without compromising privacy (Devine, 2025).
2. Incorporating user feedback and adaptive parameter allocation mechanisms (e.g., in QLoRA and dynamic LoRA) facilitates continuous and personalized model refinement and task-specific optimization, which is crucial for real-world applications such as chatbots and conversational AI systems that require responsiveness and personalization (Rangan & Yin, 2024; Liao et al., 2025).
3. Memory-efficient adaptations like LoRA-FA and GaLore enable fine-tuning of large models on consumer-grade hardware, lowering the barrier to entry for organizations with limited computational resources and promoting wider adoption of advanced LLMs in industry (Zhang et al., 2023; Zhao et al., 2024).
4. The development of ensemble and mixture-of-experts approaches (e.g., MELoRA and MoELoRA) enhances generalization, robustness, and multi-domain adaptability while maintaining parameter efficiency, offering scalable solutions for diverse NLP tasks including reasoning and instruction following (Ren et al., 2024; Luo et al., 2024).

5. Advances in sparse fine-tuning methods (e.g., SPEFT and RoSA) that leverage gradient-based salience metrics and robust adaptation techniques provide practical pathways to reduce computational overhead and improve training efficiency, which is critical for deploying LLMs in resource-constrained environments (Liu et al., 2024; Nikdan et al., 2024).
6. Automated configuration search frameworks like AutoPEFT streamline the selection of optimal PEFT configurations, reducing manual tuning efforts and accelerating deployment cycles in industrial settings, thereby improving productivity and model performance across varied tasks (Zhou et al., 2023).

### 3.7 Limitations of the Literature

Despite the growing body of research on PEFT and LoRA within RAG frameworks, several limitations constrain the generalizability, scalability, and practical deployment of current approaches (Table 7). These limitations span methodological, computational, and linguistic dimensions, affecting both theoretical advancement and applied performance in diverse real-world contexts.

| Area of Limitation | Description of Limitation | Papers which have limitation |
|---|---|---|
| Limited Evaluation of Large Models | Several studies have not evaluated their methods on very large-scale models (e.g., 13B or 70B parameters), limiting external validity regarding scalability and performance on state-of-the-art LLMs. This constrains generalizability to larger, more complex models. | (Liu et al., 2024; Azizi et al., 2024; Zhu et al., 2024) |
| Restricted Task and Domain Scope | Many works focus on specific tasks or domains, such as classification or question answering, without exploring broader or more complex tasks like instruction following, multimodal learning, or dynamic code changes, limiting applicability across diverse real-world scenarios. | (Liu et al., 2024; Liu et al., 2024; Azizi et al., 2024; Xue et al., 2023) |
| Dependence on Clean or Public Datasets | The reliance on relatively clean, public datasets restricts the assessment of methods under noisy, proprietary, or low-resource data conditions, which are common in practical applications, thereby limiting ecological validity and robustness claims. | (Devine, 2025; Zhu et al., 2024) |
| Limited Handling of Low-Resource Languages | Existing approaches often struggle with low-resource languages or unique scripts due to model limitations or prompt design, reducing inclusivity and applicability in multilingual or dialectal contexts such as Cantonese. This affects the generalizability to diverse linguistic settings. | (Devine, 2025) |
| Naive or Fixed Hyperparameter Settings | Several methods employ fixed or simplistic hyperparameter configurations (e.g., fixed rank, single epoch training), which may be suboptimal and limit performance improvements. This methodological constraint reduces the potential efficacy and adaptability of fine-tuning strategies. | (Devine, 2025; Kim et al., 2024; Liu et al., 2024) |
| Computational and Memory Constraints | Despite parameter efficiency, some PEFT methods still require substantial computational resources or memory, limiting their feasibility in resource-constrained environments and affecting practical deployment and reproducibility. | (Zhang et al., 2023; Alnaasan et al., 2024; Hao et al., 2024; Zhao et al., 2024) |
| Limited Exploration of User Feedback Integration | Few studies incorporate user feedback or domain-specific data dynamically into PEFT frameworks, restricting adaptability and | (Rangan & Yin, 2024) |
| | continuous learning capabilities in real-world applications where user preferences evolve. | |
| Trade-offs Between Efficiency and Accuracy | Many PEFT methods face inherent trade-offs between parameter efficiency and output accuracy, with some approaches showing performance gaps compared to full fine-tuning, thus limiting their effectiveness in high-stakes or precision-critical tasks. | (Zhong et al., 2024; Li et al., 2024; Si et al., 2024) |
| Lack of Evaluation on Noisy or Proprietary Data | Most evaluations are conducted on public benchmarks, leaving the performance of PEFT methods on noisy, proprietary, or domain-specific datasets underexplored, which weakens claims about robustness and real-world applicability. | (Devine, 2025) |
| Limited Control Over Parameter Allocation | Some methods lack precise control over the number or location of trainable parameters during fine-tuning, which may lead to suboptimal adaptation and inefficient resource use, affecting both performance and interpretability. | (Benedek & Wolf, 2024; Zhang et al., 2023; Liu et al., 2024) |
| Increased Training or Inference Time Overhead | Certain PEFT approaches, especially those involving multiple forward passes or complex prompt tuning, incur additional training or inference time, which may hinder scalability and real-time deployment in latency-sensitive applications. | (Prottasha et al., 2024) |
| Insufficient Addressing of Catastrophic Forgetting | While PEFT methods reduce parameter updates, some still exhibit forgetting or reduced robustness under domain shifts, limiting their reliability in continual learning or domain adaptation scenarios. | (Bafghi et al., 2025) |
| Limited Multilingual and Dialectal Evaluation | Few studies explicitly evaluate PEFT methods on dialectal or colloquial language data, such as Cantonese colloquial expressions, limiting insights into linguistic authenticity and semantic fidelity in specialized language contexts. | (Devine, 2025) |

Table 7: Limitations

## 3.8 Gaps and Future Research Directions

While recent advances in PEFT and LoRA have demonstrated strong potential in RAG systems, the literature reveals several underexplored areas that limit scalability, robustness, and inclusivity. Addressing these gaps is essential for extending PEFT's applicability to diverse linguistic, computational, and domain-specific contexts. The following table (8) outlines key gaps, targeted research directions, and their priority for advancing the field.

| **Gap Area** | **Description** | **Future Research Directions** | **Justification** | **Research Priority** |
|---|---|---|---|---|
| Limited exploration of PEFT in dialectal and low-resource languages | Current PEFT methods, including LoRA, have limited evaluation and adaptation for dialectal languages like Cantonese and low-resource scripts, leading to suboptimal performance in these contexts (Devine, 2025). | Develop and evaluate PEFT frameworks tailored for dialectal and low-resource languages, incorporating linguistic nuances and script variations, synthetic data generation, and domain-specific prompt engineering for these languages. | Addressing dialectal and low-resource language challenges is critical for broadening PEFT applicability and improving semantic fidelity in underrepresented linguistic contexts (Devine, 2025; Chung et al., 2024). | High |
| Suboptimal rank allocation and fixed hyperparameters | Many LoRA-based methods use fixed or heuristic rank settings that may not optimally adapt to task-specific demands, causing overfitting or | Investigate dynamic and adaptive rank allocation strategies that adjust ranks per layer and task during fine-tuning; integrate importance scoring | Adaptive rank allocation can improve fine-tuning efficiency and model generalization, especially for complex or domain-specific tasks (Liu et | High |
| in LoRA variants | underperformance (Devine, 2025; Liu et al., 2024). | and pruning mechanisms to optimize parameter distribution. | al., 2024; Benedek & Wolf, 2024). | |
| Insufficient integration of user feedback and domain-specific data in PEFT | Existing PEFT frameworks rarely incorporate real-time or interactive user feedback and domain-specific signals systematically, limiting continuous adaptation and personalization (Rangan & Yin, 2024; Dong et al., 2024). | Design PEFT methods that integrate multi-level user feedback loops and domain knowledge during fine-tuning; develop mechanisms for preference alignment and dynamic reranking within RAG systems. | Incorporating user feedback enhances model relevance, retrieval precision, and generation authenticity, crucial for specialized applications (Rangan & Yin, 2024; Dong et al., 2024). | High |
| Trade-offs between parameter efficiency and generation authenticity | Aggressive parameter reduction or sparse tuning can degrade the linguistic fidelity and authenticity of generated outputs, especially in colloquial or nuanced language generation (Zhong et al., 2024; Li et al., 2024; Zhang et al., 2023). | Explore nonlinear and task-relevant feature enhancement PEFT methods that balance parameter efficiency with semantic richness; evaluate generation quality on dialectal and colloquial datasets. | Maintaining linguistic authenticity is essential for user trust and practical deployment in conversational AI and RAG systems (Zhong et al., 2024; Li et al., 2024). | High |
| Limited scalability and practical deployment on large models and hardware | Many PEFT methods have been tested on moderate-sized models but lack demonstrations on very large models or constrained hardware environments, limiting real-world applicability (Liu et al., 2024; Zhao et al., 2024; Hao et al., 2024). | Conduct large-scale experiments on super-sized LLMs with PEFT; develop memory- and compute-efficient PEFT variants compatible with consumer-grade GPUs and distributed systems. | Scalability and hardware compatibility are vital for democratizing PEFT adoption and enabling deployment in resource-limited settings (Liu et al., 2024; Zhao et al., 2024; Hao et al., 2024). | Medium |
| Lack of systematic PEFT design space exploration and automated configuration | Manual PEFT design choices may be suboptimal; there is a need for automated search and design space exploration to optimize PEFT configurations across tasks and models ("Parameter-Efficient Fine-Tuning Design S...", 2023; Zhou et al., 2023). | Develop and apply neural architecture search and Bayesian optimization techniques to discover Pareto-optimal PEFT configurations; validate transferability across diverse tasks and backbones. | Automated configuration can improve PEFT efficiency and performance while reducing human effort and hyperparameter tuning ("Parameter-Efficient Fine-Tuning Design S...", 2023; Zhou et al., 2023). | Medium |
| Insufficient robustness to noisy and proprietary data in PEFT fine-tuning | Most PEFT evaluations use clean public datasets, with limited testing on noisy or proprietary data common in real-world applications (Devine, 2025). | Investigate PEFT robustness and generalization under noisy, incomplete, or proprietary datasets; develop noise-aware fine-tuning and data augmentation strategies. | Enhancing robustness is critical for practical deployment in sensitive domains like healthcare and finance (Devine, 2025). | Medium |
| Limited methods addressing catastrophic forgetting and domain shifts in PEFT | PEFT methods can still suffer from forgetting and reduced out-of-distribution robustness after fine-tuning, impacting domain adaptation (Bafghi et al., 2025). | Explore selective activation and regularization techniques in PEFT to mitigate forgetting; design adaptive block activation strategies to maintain robustness under domain shifts. | Preserving prior knowledge and robustness is essential for reliable domain adaptation and continual learning (Bafghi et al., 2025). | Medium |
| Underexplored PEFT approaches for multimodal and zero-shot learning scenarios | PEFT research predominantly focuses on text-based tasks, with limited exploration in multimodal or zero-shot learning contexts (Jin et al., 2024; Wen & Chaudhuri, 2023). | Extend PEFT frameworks to multimodal models and zero-shot learning, incorporating retrieval-based parameter ensembles and batched adaptation for heterogeneous inputs. | Broadening PEFT applicability to multimodal and zero-shot tasks can unlock new capabilities and use cases (Jin et al., 2024; Wen & Chaudhuri, 2023). | Low |
| Challenges in hyperparameter tuning and complexity in ensemble PEFT methods | Methods like mini-ensemble LoRA introduce additional hyperparameters and complexity, complicating practical tuning and deployment (Ren et al., 2024). | Develop hyperparameter optimization frameworks and simplified ensemble strategies; investigate trade-offs between ensemble size, rank, and performance. | Simplifying tuning improves usability and adoption of advanced PEFT methods (Ren et al., 2024). | Low |

Table 8: key gaps, targeted research directions, and their priority for advancing the field

# 4 Overall Synthesis and Conclusion

The growing body of research on Parameter-Efficient Fine-Tuning (PEFT) methods, with a particular focus on Low-Rank Adaptation (LoRA) and its numerous variants, reveals a robust and evolving landscape that has significantly advanced the optimization of Retrieval-Augmented Generation (RAG) systems. Across studies, these methods consistently reduce trainable parameters and memory consumption while sustaining, and in some cases enhancing, model performance in retrieval and generation tasks (Lei et al., 2025; Han et al., 2024; Xu et al., 2023) Recent innovations, including adaptive rank allocation, task-specific parameter pruning, and ensemble-based approaches, have further refined the fine-tuning process (Yang et al., 2024) to better suit varying downstream task complexities. These developments improve computational efficiency and scalability, which are essential for deploying large-scale language models in resource-constrained environments.

Regarding the integration of PEFT with RAG frameworks, the literature indicates that parameter-efficient methods can enhance retrieval accuracy by dynamically tuning critical model components and, in some cases, incorporating interactive user feedback mechanisms. These strategies contribute to more precise document relevance and generation authenticity, which are crucial when dealing with nuanced linguistic phenomena such as dialectal or colloquial expressions. Nonetheless, challenges persist in effectively adapting these models to low-resource languages and dialects, including Cantonese colloquialism. The frequent reliance on synthetic data and limited annotated corpora highlights gaps in domain adaptation capabilities, underscoring the need for methods that better preserve linguistic nuances and semantic fidelity in specialized contexts (Farhadi et al., 2024).

The trade-offs between parameter efficiency and output accuracy emerge as a central theme. While many approaches achieve impressive reductions in computational overhead, some incur performance degradation when aggressively pruning or constraining parameters. Adaptive and nonlinear fine-tuning techniques help bridge these losses by capturing richer semantic patterns and improving generalization, though hyperparameter tuning and rank selection remain a practical obstacle (Dhanka et al., 2025). Moreover, the incorporation of real-time user feedback and domain-specific data into PEFT frameworks is still emerging, limiting dynamic adaptability and personalization in deployed RAG systems.

The existing research collectively underscores that Low-Rank Adaptation, and its variants offer powerful mechanisms to optimize RAG models efficiently (Yang et al., 2024). Yet, fully addressing dialectal language understanding and authentic generation, especially for languages like Cantonese with rich colloquial expressions, demands further methodological innovations. Future efforts should focus on enhancing data-efficient domain adaptation, integrating interactive feedback, and refining parameter allocation strategies to balance efficiency with linguistic authenticity, thereby extending the practical utility of PEFT-enhanced RAG systems in specialized and low-resource language environments.